%% file: 2013_LPNMR_EpistemicPlanning.tex
\documentclass{llncs}

\usepackage{times}
\usepackage[table]{xcolor}%

\input{macros.tex}

\usepackage[numbers]{natbib}

\usepackage{balance}
\usepackage{trfsigns}

\newtheorem{Theorem}{Theorem}
\newtheorem{Example}{Example}
\newtheorem{Conjecture}{Conjecture}
\newtheorem{Definition}{Definition}
\newtheorem{Observation}{Observation}

\begin{document}

\title{\emph{\huge h-approximation}:\\History-Based Approximation to Possible World Semantics as ASP}
%
%
\author{Manfred Eppe \and Mehul Bhatt \and Frank Dylla}
%
%
\institute{University of Bremen, Germany\\
\email{\color{blue!80!black}\{meppe,bhatt,dylla\} @ informatik.uni-bremen.de}\\ 
}

\maketitle

\begin{abstract}\footnotesize
 We propose a history-based approximation of the Possible Worlds Semantics  ($\mathcal{PWS}$) for reasoning about knowledge and action. A respective planning system is implemented by a transformation of the problem domain to an Answer-Set Program. The novelty of our approach is elaboration tolerant support for postdiction under the condition that the plan existence problem is still solvable in NP, as compared to $\Sigma_2^P$ for non-approximated \pws\ of \citet{Son2001}. We demonstrate our planner with standard problems and present its integration in a cognitive robotics framework for high-level control in a smart home. 
\end{abstract}

\section{Introduction}
\vspace{-9pt}
Dealing with incomplete knowledge in the presence of abnormalities, unobservable processes, and other real world considerations is a crucial requirement for real-world planning systems.  
Action-theoretic formalizations for handling incomplete knowledge can be traced back to the Possible Worlds Semantics (\pws) of \citet{Moore1985}. 
Naive formalizations of the \pws\ result in search with complete knowledge in an exponential number of possible worlds.
The planning complexity for each of these worlds again ranges from polynomial to exponential time \citep{Backstrom1995} (depending on different assumptions and restrictions). 
\citet{Baral2000} show that in case of the action language \Ak\ the planning problem is $\Sigma_2^P$ complete (under certain restrictions).
This high complexity is a problem for the application of epistemic planning in real-world applications like cognitive robotics or smart environments, where real-time response is needed.
One approach to reduce complexity is the approximation of \pws. 
\citet{Son2001} developed the 0-approximation semantics for \Ak which results in an NP-complete solution for the plan existence problem.
However, the application of approximations does not support all kinds of epistemic reasoning, like \i{postdiction} --  a useful inference pattern of knowledge acquisition, e.g., to perform failure diagnosis and abnormality detection. 
Abnormalities are related to the \i{qualification problem}: it is not possible to model all conditions under which an action is successful.
A partial solution to this is \i{execution monitoring} (\eg \citep{Petersson2005}), \ie action success is observed by means of specific sensors.
If expected effects are not achieved, one can \i{postdict} about an occurred abnormality. 


%
%
%
%


In Section \ref{sec:ASPFormalization} we present the core contribution of this paper: a `history' based approximation of the \pws\ --- called  \i{h-approximation} (\hpx) --- which supports postdiction. 
Here, the notion of history is used in an epistemic sense of maintaining and refining knowledge about the past by postdiction and commonsense law of inertia.
For instance, if an agent moves trough a door (say at $t=2$) and later (at some $t'>2$) comes to know that it is behind the door, then it can postdict that the door must have been open at $t=2$. 
Solving the plan-existence problem with h-approximation is in NP and finding optimal plans is in $\Delta_2^P$.
Despite the low complexity of \hpx\ compared to \Ak
\footnote{Throughout the paper we usually refer to the full \pws\ semantics of \Ak. Whenever referring to the 0-approximation semantics this is explicitly stated.}
 it is more expressive in the sense that it allows to make propositions about the past.
Hence, the relation between \hpx\ and \Ak\ is not trivial and deserves a thorough investigation which is provided in Section \ref{sec:relToPWS}:
We extend \Ak\ and define a \i{temporal query semantics} (\Ak$^{TQS}$) which allows to express knowledge about the past.
This allows us to show that \hpx\ is sound \wrt a temporal possible worlds formalization of action and knowledge.

A planning system for \hpx\ is developed via its interpretation as an Answer Set Program (ASP).
The formalization supports both sequential and (with some restrictions) concurrent planning, and 
\emph{conditional plans} are generated with off-the-shelf ASP solvers. 
We provide a case study in a smart home as a proof of concept in Section \ref{sec:evalAndCasStudy}.
\vspace{-9pt}
\section{Related Work}
\vspace{-11pt}
\label{sec:relWork}
Approximations of the \pws\ have been proposed, primarily driven by the need to reduce the complexity of planning with incomplete knowledge vis-a-vis the tradeoff with support for expressiveness and inference capabilities. 
For such approximations, we are interested in: (i) the extent to which \emph{postdiction} is supported; (ii) whether they are \i{guaranteed to be epistemically accurate}, (iii) their \i{tolerance to problem elaboration} \citep{Mccarthy1998} and (iv) their \i{computational complexity}. 
We identified that many approaches indeed support postdiction, but only in an ad-hoc manner: Domain-dependent postdiction rules and knowledge-level effects of actions are implemented manually and depend on correctness of the manual encoding. 
For this reason, epistemic accuracy is not guaranteed.
Further, even if postdiction rules are implemented epistemically correct \wrt a certain problem, then correctness of these rules may not hold anymore if the problem is elaborated (see Example \ref{ex:AckNotElTolerant}): Hence, ad-hoc formalization of postdiction rules is not elaboration tolerant. 

\f{Epistemic Action Formalisms.} \quad 
\citet{Scherl2003} provide an epistemic extension and a solution to the frame problem for the Situation Calculus (SC)
, and \citet{Patkos2009} as well as \citet{Miller2013} provide epistemic theories for the Event Calculus.
These approaches are all complete \wrt \pws\ and hence suffer from a high computational complexity. 
%
%
%
\citet{Thielscher2000} describes how knowledge is represented in the Fluent Calculus (FC). 
The implementation in the FC-based framework FLUX is not elaboration-tolerant as it requires manual encoding of knowledge-level effects of actions.
\citet{Liu2005} use a progression operator to approximate \pws . 
The result is a tractable treatment of the projection problem, but again postdiction is not supported.
The PKS planner \citep{Petrick2004} is able to deal with incomplete knowledge, but postdiction is only supported in an ad-hoc manner.
%
\citet{Vlaeminck2012} propose a first order logical framework to approximate \pws.
The framework supports reasoning about the past, allows for elaboration tolerant postdiction reasoning, and the projection problem is solvable in polynomial time when using their approximation method. 
However, the authors do not provide a practical implementation and evaluation and they do not formally relate their approach to other epistemic action languages. 
To the best of our knowledge, besides \cite{Vlaeminck2012,Miller2013} there exists no approach which employs a postdiction mechanism that is based on explicit knowledge about the past.

\noindent There exist several PDDL-based planners that deal with incomplete knowledge. 
These planners typically employ some form of \pws\ semantics and achieve high performance via practical optimizations such as BDDs \citep{Cimatti2003} or heuristics that build on a relaxed version of the planning problem \citep{Hoffmann2005}. 
The way how states are modeled can also heavily affect performance, as shown by \citet{To2011} with the \i{minimal-DNF} approach. 
With \hpx, we propose another alternative state representation which is based on explicit knowledge about the past. 
%
%

\f{The \mc{A}-Family of Languages.}\quad
The action language \A\ \citep{Gelfond1993} is originally defined for domains with complete knowledge. Later, epistemic extensions which consider incomplete knowledge and sensing were defined. 
Our work is strongly influenced by these approaches \citep{Lobo2001,Son2001,Tu2007}: \quad
\citet{Lobo2001} use epistemic logic programming and formulate a \pws\ based epistemic semantics.
The original \Ak\ semantics is based on \pws\ and (under some restrictions) is sound and complete \wrt the approaches by \citet{Lobo2001} and \citet{Scherl2003}.
\citet{Tu2007} introduce \Ack\ and add Static Causal Laws (SCL) to the 0-approximation semantics of \Ak.
They implement \Ack\ in form of the ASCP planning system which -- like \hpx\ -- is based on ASP. 
The plan-existence problem for \Ack\ is still NP-complete \citep{Tu2007}.
%
The authors demonstrate that SCL can be used for an ad-hoc implementation of postdiction. 
However, we provide the following example to show that an ad-hoc realisation of postdiction is not \i{elaboration tolerant}:
%
\begin{Example}\label{ex:AckNotElTolerant}
{\small
\begin{sloppypar}
A robot can drive into a room through a door $d$. 
It will be in the room if the door is open: \f{causes}(\cos{drive$_d$},\cos{in},\{\cos{open$_d$}\}).
An auxiliary fluent \cos{did\_drive$_d$} represents that the action has been executed: \f{causes}(\cos{drive$_d$},\cos{did\_drive$_d$},$\emptyset$); 
A manually encoded SCL \f{if}(\cos{open$_d$},\{\cos{did\_drive$_d$},\cos{in}\}) postdicts that if the robot is in the destination room after driving the door must be open. 
The robot has a location sensor to determine whether it arrived: \f{determines}(\cos{sense\_in},\cos{in}).
Consider an empty initial state $\delta_{init} = \emptyset$, a door $d=1$ and a sequence $\alpha=[\cos{drive}_1;\cos{sense\_in}]$. 
Here \Ack\ correctly generates a state $\delta' \supseteq \{\cos{open}_1\}$ where the door is open if the robot is in the room. 
Now consider an elaboration of the problem with two doors ($d \in \{1,2\}$) and a sequence $\alpha=[\cos{drive}_1;\cos{drive}_2;\cos{sense\_in}]$. 
By Definitions 4--8 and the closure operator $CL_D$ in \cite{Tu2007}, \Ack\ produces a state $\delta'' \supseteq \{\cos{open}_1,\cos{open}_2\}$ where the agent knows that door 1 is open, even though it may actually be closed: this is not sound \wrt \pws\ semantics.
\end{sloppypar}
}
\end{Example}
Another issue is \i{concurrent acting and sensing}. 
\citet{Son2001} (p. 39) describe a modified transition function for the \emph{0-approximation} to support this form of concurrency: they model sensing as determining the value of a fluent after the physical effects are applied. 
However, this workaround does not support some trivial commonsense inference patterns: 
\vspace{-3pt}
\begin{Example}\label{ex:YaleShootingMod}
{\small
Consider a variation of the Yale shooting scenario where an agent can sense whether the gun was loaded when pulling the trigger because she hears the bang. 
Without knowing whether the gun was initially loaded, the agent should be able to immediately infer whether or not the turkey is dead depending on the noise.
This is not possible with the proposed workaround because it models sensing as the acquisition of a fluent's value after the execution of the sensing: Here the gun is unloaded after executing the shooting, regardless of whether it was loaded before. 
\hpx\ allows for such inference because here sensing yields knowledge about the value of a fluent at the time sensing is executed.
}
\end{Example}
%
\section{h-approximation and its Translation to ASP}\label{sec:ASPFormalization}
\vspace{-8pt}
The formalization is based on a foundational theory $\Gamma_{hapx}$ and on a set of \emph{translation rules} \f{T} that are applied to a planning domain \mc{P}. 
\mc{P} is modelled using a PDDL like syntax and consists of the language elements in (\ref{eq:syntax:vp}-\ref{eq:syntax:ep}) as follows: 
Value propositions ($\mathcal{VP}$) denote initial facts (\ref{eq:syntax:vp}); 
Oneof constraints ($\mathcal{OC}$) denote exclusive-or knowledge (\ref{eq:syntax:oc}); 
Goal propositions ($\mathcal{G}$) denote goals%
\footnote{$\sn{type}$ is either \co{weak} or \co{strong}. A weak goal must be achieved in only one branch of the conditional plan while a strong goal must be achieved in all branches (see \eg \citep{Cimatti2003}).}
 (\ref{eq:syntax:gp});
Knowledge propositions ($\mathcal{KP}$) denote sensing (\ref{eq:syntax:kp}); 
Executability conditions ($\mathcal{EXC}$) denote what an agent must know in order to execute an action (\ref{eq:syntax:exp});
Effect propositions ($\mathcal{EP}$) denote conditional action effects (\ref{eq:syntax:ep}).
\smallskip
{\small
\begin{subequations}\label{eq:syntax:all}
\setlength\abovedisplayskip{-0pt}
\setlength\belowdisplayskip{-7pt}
\begin{minipage}[l]{0.25\textwidth}
\begin{equation}\label{eq:syntax:vp}
\begin{aligned}
&\cos{(:init $l^{init}$)}
\end{aligned}
\end{equation}
\end{minipage}
\begin{minipage}[l]{0.33\textwidth}
\begin{equation}\label{eq:syntax:oc}
\begin{aligned}
&\cos{(oneof $l^{oo}_1 \hdots l^{oo}_n$)}
\end{aligned}
\end{equation}
\end{minipage}
\begin{minipage}[l]{0.40\textwidth}
\begin{equation}\label{eq:syntax:gp}
\begin{aligned}
&\cos{(:goal} \sns{ type } \cos{(and $l^{g}_1 \hdots l^{g}_n$))}
\end{aligned}
\end{equation}
\end{minipage}
%

\begin{minipage}[l]{0.25\textwidth}
\begin{equation}\label{eq:syntax:kp}
\begin{aligned}
&\cos{(:action $a$}\\[-5pt]
&\cos{ :observe $f$)}
\end{aligned}
\end{equation}
\end{minipage}
\begin{minipage}[l]{0.33\textwidth}
\begin{equation}\label{eq:syntax:exp}
\begin{aligned}
&\cos{(:action $a$ executable}\\[-5pt]
&~~~~~~~~\cos{(and$\,\,l^{ex}_1\hdots l^{ex}_n$))}
\end{aligned}
\end{equation}
\end{minipage}
\begin{minipage}[l]{0.40\textwidth}
\begin{equation}\label{eq:syntax:ep}
\begin{aligned}
&\cos{(:action $a$} \cos{ :effect}\\[-5pt]
&\cos{ when (and $l^c_1 \hdots l^c_n$) $l^e$)}
\end{aligned}
\end{equation}
\end{minipage}
\end{subequations}
}

\smallskip
Formally, a planning domain \mc{P} is a tuple {\small $\rab{\mc{I}, \mc{A}, \mc{G}}$} where:
\vspace{-3pt}
\begin{itemize}
{
	\item \mc{I} is a set of value propositions (\ref{eq:syntax:vp}) and oneof-constraints (\ref{eq:syntax:oc})
	\item \mc{A} is a set of actions. An action $a$ is a tuple {\small $\rab{\mc{EP}^a, \mc{KP}^a, \mc{EXC}^a}$} consisting of a set of effect propositions $\mc{EP}^a$ (\ref{eq:syntax:ep}), a set of knowledge propositions $\mc{KP}^a$ (\ref{eq:syntax:kp}) and an executability condition $\mc{EXC}^a$ (\ref{eq:syntax:exp}).
	\item \mc{G} is a set of goal propositions (\ref{eq:syntax:gp}).
}	
\end{itemize}
\vspace{-3pt}
An ASP translation of \mc{P}, denoted by LP(\mc{P}), consists of a domain-dependent theory and a domain-independent theory:
\vspace{-3pt}
\begin{itemize}
{
	\item Domain-dependent theory ($\Gamma_{world}$): It consists of a set of rules $\Gamma_{ini}$ representing initial knowledge; $\Gamma_{act}$ representing actions; and $\Gamma_{goals}$ representing goals. 
	\item Domain-independent theory ($\Gamma_{hapx}$): This consists of a set of rules to handle inertia ($\Gamma_{in}$); sensing ($\Gamma_{sen}$);  concurrency ($\Gamma_{conc}$), plan verification ($\Gamma_{verify}$) as well as plan-generation \& optimization ($\Gamma_{plan}$).
}

\end{itemize}
\vspace{-5pt}
The resulting Logic Program LP(\mc{P}) is given as:
\vspace{-6pt}
\begin{align}
LP(\mc{P}) = ~[~\Gamma_{in}&\cup\Gamma_{sen}\cup\Gamma_{conc}\cup\Gamma_{verify}\cup\Gamma_{plan}~] \cup
							[~\Gamma_{ini}\cup\Gamma_{act}\cup\Gamma_{goal}] \\[-23pt]\notag
\end{align}
\vspace{-6pt}
%
\begin{center}
\fcolorbox{black}{yellow!0!white}{\parbox{0.97\textwidth}{
\noindent {\small 
\textbf{Notation}.\quad We use the variable symbols \co{A} for \emph{action}, \co{EP} for \emph{effect proposition}, \co{KP} for \emph{knowledge proposition}, \co{T} for \emph{time} (or step), \co{BR} for \emph{branch}, and \co{F} for \emph{fluent}. \co{L} denotes \emph{fluent literals} of the form \co{F} or $\neg$\co{F}. $\ol{\co{L}}$ denotes the complement of \co{L}. For a predicate \co{p($\hdots$,L,$\hdots$)} with a literal argument, we denote strong negation ``$-$'' with the $\neg$ symbol as prefix to the fluent.  For instance, we denote \co{-knows(F,T,T,BR)} by \co{knows($\neg$ F,T,T,BR)}. 
$|\co{L}|$ is used to ``positify'' a literal, \ie $|\neg\co{F}| = \co{F}$ and $|\co{F}| = \co{F}$. 
Respective small letter symbols denote constants. 
For example \co{knows($l$,$t$,$t'$,$br$)} denotes that at step $t'$ in branch $br$ it is known that literal $l$ holds at step $t$. 
}}}
\end{center}
\vspace{-18pt}
\subsection{Translation Rules:\quad (\mc{P}  {\small$\stackrel{\f{T1--T8}}{\longmapsto} \Gamma_{world}$})}
\vspace{-6pt}
\setcounter{equation}{0}
\makeatletter 
\renewcommand{\theequation}{T\@arabic\c@equation} 

The domain dependent theory $\Gamma_{world}$ is obtained by applying the set of translation rules $\f{T}=\{ \ref{lst:lp-declaration},\hdots,\ref{lst:lp-goals}\}$ on a planning domain \mc{P}.

\smallskip

\noindent  \f{Actions / Fluents Declarations \scriptsize{(T1)}.}\quad For every fluent $f$ or action $a$, LP(\mc{P}) contains:
\begin{equation}
\setlength\abovedisplayskip{-1pt}
\setlength\belowdisplayskip{-1pt}
\label{lst:lp-declaration}
fluent(f). \ms action(a). 
\end{equation}

%
\noindent\f{Knowledge ($\mc{I} \stackrel{\f{T2--T3}}{\longmapsto}\Gamma_{ini}$)}.\quad 
Facts $\Gamma_{ini}$ for initial knowledge are obtained by applying translation rules (\ref{lst:lp-initKnowledge}-\ref{lst:lp-oneof}).
For each value proposition (\ref{eq:syntax:vp}) we generate the fact:
\vspace{-5pt}
\begin{align}
\label{lst:lp-initKnowledge}
knows(l^{init},0,0,0).\\[-28pt]\notag 
\end{align}
\newpage
For each oneof-constraint (\ref{eq:syntax:oc}) with the set of literals $\f{C}=\{l^{oc}_1 \hdots l^{oc}_{n}\}$ we consider one literal $l^{oc}_i \in \f{C}$. 
Let $\{l^+_{i_1},\hdots,l^+_{i_{m}}\} = \f{C} \backslash l^{oc}_i$ be the subset of literals except $l^{oc}_i$. 
Then, for each $l^{oc}_i \in \f{C}$ we generate the LP rule:
\begin{subequations}\label{lst:lp-oneof}
\setlength\abovedisplayskip{-0pt}
\setlength\belowdisplayskip{-0pt}
\begin{equation}\label{lst:lp-oneof1}
\begin{aligned}
kno&ws( l^{oc}_i ,0,T,BR)   \lam 
&knows( \ol{l^+_{i_1}} ,0,T,BR), \hdots ,knows( \ol{l^+_{i_{m}}} ,0,T,BR).
\end{aligned}
\end{equation}
\begin{equation}\label{lst:lp-oneof2} 
\begin{aligned}
knows&( \ol{l^+_{i_1}} ,0,T,BR) \lam knows( l^{oc}_i ,0,T,BR).~~~~~~\hdots \\[-3pt]
knows&( \ol{l^+_{i_m}} ,0,T,BR) \lam knows( l^{oc}_i ,0,T,BR). 
\end{aligned}
\end{equation}
\end{subequations}
(\ref{lst:lp-oneof1}) denotes that if all literals except one are known not to hold, then the remaining one must hold.
Rules (\ref{lst:lp-oneof2}) represent that if one literal is known to hold, then all others do not hold.
At this stage of our work we only support static causal laws (SCL) to constrain the initial state, because this is the only state in which they do not interfere with the postdiction rules. 

\noindent  \f{Actions ($\mc{A} \stackrel{\f{T4--T7}}{\longmapsto}\Gamma_{act}$)}.\quad 
The generation of rules representing actions covers executability conditions, knowledge-level effects, and knowledge propositions.

\noindent \emph{Executability Conditions}.\quad  These reflect what an agent must know to execute an action. 
Let $\mc{EXC}^a$ of the form (\ref{eq:syntax:exp})  be the executability condition of action $a$ in \mc{P}. Then LP(\mc{P}) contains the following constraints, where an atom \co{occ($a$,$t$,$br$)} denotes the occurrence of action $a$ at step $t$ in branch $br$:
{
\addtolength{\abovedisplayskip}{-5pt}
\addtolength{\belowdisplayskip}{-5pt}
\begin{equation}
\label{lst:lp-actionExcondition}
\begin{aligned}
\lam occ( a ,T,BR), not \ms knows( l^{ex}_{1} ,T,T,BR). & ~~~~~~\hdots \\[-2pt]
\lam occ( a ,T,BR), not \ms knows( l^{ex}_{n} ,T,T,BR). &
\end{aligned}
\end{equation}
}
\noindent \emph{Effect Propositions}.\quad 
For every effect proposition $ep \in \mc{EP}^a$, of the form 
\co{(when (and $f^{c}_1  \hdots f^{c}_{np} \ms \neg f^{c}_{np+1}  \hdots \neg f^{c}_{nn}) \ms l^e$)}, 
LP(\mc{P}) contains (\ref{lst:lp-actionEffProp}), where \co{hasPC/2} (resp. \co{hasNC/2}) represents postive (resp. negative) condition literals, \co{hasEff/2} represents effect literals and \co{hasEP/2} assigns an effect proposition to an action:

{
\addtolength{\abovedisplayskip}{-5pt}
\addtolength{\belowdisplayskip}{-10pt}
\begin{equation}
\label{lst:lp-actionEffProp}
\begin{aligned}
&hasEP( a , ep ).~~~						  hasEff( ep , l^e ).\\[-1pt]
&hasPC( ep , f^{c}_1 ).~			 \hdots   hasPC(ep, f^{c}_{np} ).~  \hdots\\
&~~~~~~~~~hasNC( ep , f^{c}_{np+1} ).~	 \hdots  hasNC(ep, f^{c}_{nn} ).
\end{aligned}
\end{equation}
}

\noindent \emph{Knowledge Level Effects of Non-Sensing Actions}.\quad 
(\ref{lst:lp-effCondTrans1}-\ref{lst:lp-effCondTrans3})%
\footnote{The frame problem is handled by minimization in the stable model semantics (see \eg \citep{Lee2012a}).}
{
\begin{subequations}\label{eq:trans-three-rules}\small
\begin{equation}\label{lst:lp-effCondTrans1}
\begin{aligned}
knows(l^e,T+1,T1,BR) \lam &apply(ep,T,BR),T1>T,\\[-1pt]
&knows(l^c_1,T,T1,BR),\hdots,knows(l^c_n,T,T1,BR).
\end{aligned}
\end{equation}
\vspace{-6pt}
\begin{equation}\label{lst:lp-effCondTrans2}
\begin{aligned}
knows( l^c_i ,T,T1,BR) \lam &apply(ep ,T,BR),\\[-1pt]
&knows( l^e ,T+1,T1,BR), knows( \ol{l^e} ,T,T1,BR).
\end{aligned}
\end{equation}
\vspace{-6pt}
\begin{equation}\label{lst:lp-effCondTrans3}
\begin{aligned}
knows( \ol{l^{c-}_i} ,T,T1,BR) \lam& apply(ep ,T,BR),knows( \ol{l^e} ,T+1,T1,BR),\\[-1pt]
	&knows( l^{c+}_{i_1} ,T,T1,BR), \hdots ,knows( l^{c+}_{i_n} ,T,T1,BR).\\
\end{aligned}
\end{equation}
\end{subequations}
}

$\blacktriangleright$ \emph{Causation} (\ref{lst:lp-effCondTrans1}).\quad If all condition literals $l^c_i$ of an EP (\ref{eq:syntax:ep}) are known to hold at $t$, and if the action is applied at $t$, then at $t' > t$, it is known that its effects hold at $t+1$. The atom \co{apply($ep$,$t$,$br$)} represents that $a$ with the EP $ep$ happens at $t$ in $br$.

\smallskip

$\blacktriangleright$ \emph{Positive postdiction} (\ref{lst:lp-effCondTrans2}).\quad For each condition literal $l^c_i \in \{l^c_1, \hdots, l^c_k\}$ of an effect proposition $ep$ we add a rule (\ref{lst:lp-effCondTrans2}) to the LP. 
This defines how knowledge about the condition of an effect proposition is postdicted by knowing that the effect holds after the action but did not hold before. For example, if at $t'$ in $br$ it is known that the complement $\ol{l}$ of an effect literal of an EP holds at some $t<t'$ (i.e., \co{knows($\ol{l}$,$t$,$t'$,$br$)}), and if the EP is applied at $t$, and if it is known that the effect literal holds at $t+1$ (\co{knows($l$,$t+1$,$t'$,$br$)}), then the EP must have set the effect. Therefore one can conclude that the conditions $\{l^c_1, \hdots, l^c_k\}$ of the EP must hold at $t$.

\smallskip

$\blacktriangleright$ \emph{Negative postdiction} (\ref{lst:lp-effCondTrans3}).\quad 
For each potentially unknown condition literal $l^{c-}_i \in \{l^c_1, \hdots, l^c_n\}$ of an effect proposition $ep$ we add one rule (\ref{lst:lp-effCondTrans3}) to the program, where $\{l^{c+}_{i_1},\hdots,l^{c+}_{i_n}\} = \{l^c_1, \hdots, l^c_n\} \backslash l^{c-}_i$ are the condition literals that are known to hold.
This covers the case where we postdict that a condition must be false if the effect is known not to hold after the action and all other conditions are known to hold. For example, if at $t'$ it is known that the complement of an effect literal $l$ holds at some $t+1$ with $t+1\leq t'$, and if the EP is applied at $t$, and if it is known that all condition literals hold at $t$, except one literal $l^{c-}_i$ for which it is unknown whether it holds. 
Then the complement of $l^{c-}_i$ must hold because otherwise the effect literal would hold at $t+1$. 

\smallskip

\noindent \emph{Knowledge Propositions}.\quad We assign a KP (\ref{eq:syntax:kp}) to an action $a$ using \co{hasKP/2}:
{
\addtolength{\abovedisplayskip}{-3pt}
\addtolength{\belowdisplayskip}{-12pt}
\begin{equation}\label{lst:lp-assignKPro}
\begin{aligned}
hasKP(a,f).
\end{aligned}
\end{equation}
}
\begin{sloppypar}
\noindent \f{Goals ($\mc{G} \stackrel{\f{T8}}{\longmapsto}\Gamma_{goal}$)}.\quad For literals $l^{sg}_1,...,l^{sg}_n$ in a strong goal proposition and $l^{wg}_1, ...,l^{wg}_m$ in a weak goal proposition we write:
\end{sloppypar}
{
\addtolength{\abovedisplayskip}{-5pt}
\addtolength{\belowdisplayskip}{-5pt}
\begin{subequations}\label{lst:lp-goals} 
\begin{equation}\label{lst:lp-wGoals} 
\begin{aligned}
sGoal(T,BR) \lam knows( l^{sg}_1 ,T,T,BR),  \text{...},
					knows( l^{sg}_n ,T,T,BR), s(T), br(BR).
\end{aligned}
\end{equation}
\begin{equation}\label{lst:lp-sGoals} 
\begin{aligned}
\hspace{-2pt}
wGoal(T,BR) \lam knows( l^{wg}_1 ,T,T,BR),  \text{...} , 
					knows( l^{wg}_m ,T,T,BR), s(T), br(BR).
\end{aligned}
\end{equation}
\end{subequations}
}
where an atom \co{sGoal($t$,$br$)} (resp. \co{wGoal($t$,$br$)}) represents that the strong (resp. weak) goal is achieved at $t$ in $br$.

\setcounter{equation}{2}
\makeatletter 
\renewcommand{\theequation}{\@arabic\c@equation} 
\vspace{-13pt}
\subsection{$\Gamma_{hapx}$ -- Foundational Theory (F1--F5)}
\vspace{-5pt}
\label{ssec:domIndep}
The foundational domain-independent \hpx-theory is shown in Listing \ref{lst:lpIndep}. 
It covers concurrency, inertia, sensing, goals, plan-generation and plan optimization.
Line \ref{lst:lpIndep:defStepsBranches} sets the maximal plan length \co{maxS} and width \co{maxBr}. 


\noindent  \f{F1.\quad Concurrency ($\Gamma_{conc}$)}\quad 
Line \ref{lst:lpIndep:applyEffPro} applies all effect propositions of an action $a$ if that action occurs.
We need two restrictions regarding concurrency of non-sensing actions:
effect similarity and effect contradiction.
Two effect propositions are similar if they have the same effect literal.
Two EPs are contradictory if they have complementary effect literals and if their conditions do not contradict (\li{lst:lpIndep:noContraConcurrentEnd}). 
The cardinality constraint \li{lst:lpIndep:noSimilarConcurrent} enforces that two similar EPs (with the same effect literal) do not apply concurrently, whereas \li{lst:lpIndep:noContraConcurrentStart} restricts similarly for contradictory EPs.



\noindent  \f{F2.\quad  \f{Inertia}  ($\Gamma_{in}$)}\quad Inertia is applied in both forward and backward direction similar to \citep{Gelfond1993}.
To formalize this, we need a notion on knowing that a fluent is \i{not} initiated (resp. terminated). 
This is expressed with the predicates \co{kNotInit}/\co{kNotTerm}.%
\footnote{For brevity Listing \ref{lst:lpIndep} does only contain rules for \co{{kNotInit}}; the rules for \co{kNotTerm} are analogous resp. to \lir{lst:lpIndep:epNotInitiatedStart}{lst:lpIndep:epFalseInitiated}.}
A fluent could be known to be not  initiated for two reasons: (1) if no effect proposition with the respective effect fluent is applied, then this fluent can not be initiated. \co{initApp($f$,$t$,$br$)} (\li{lst:lpIndep:epNotInitiatedStart}) represents that at $t$ an EP with the effect fluent $f$ is applied in branch $br$. 
If \co{initApp($f$,$t$,$br$)} does not hold then $f$ is known not to be initiated at $t$ in $br$ (\li{lst:lpIndep:epNotInitiatedEnd}).

%

\begin{lstlisting}[style=lp,label=lst:lpIndep,escapeinside={!@}{!@},caption=Domain independent theory ($\Gamma_{hapx}$)]
s(0..maxS). ss(0..maxS-1). br(0..maxBr). !@\label{lst:lpIndep:defStepsBranches}!@
!@$\blacktriangleright$ \f{Concurrency ($\Gamma_{conc}$)}!@ 
apply(EP,T,BR) :- hasEP(A,EP), occ(A,T,BR). !@\label{lst:lpIndep:applyEffPro}!@
contra(EP1,EP) :- hasPC(EP1,F),hasNC(EP,F). !@\label{lst:lpIndep:noContraConcurrentEnd}!@
:- 2{apply(EP,T,BR):hasEff(EP,F)},br(BR), s(T), fluent(F).!@\label{lst:lpIndep:noSimilarConcurrent}!@
:- apply(EP,T,BR), hasEff(EP,F), apply(EP1,T,BR), 			hasEff(EP1,$\neg$F), EP != EP1, not contra(EP1,EP).!@\label{lst:lpIndep:noContraConcurrentStart}!@
!@$\blacktriangleright$ \f{Inertia  ($\Gamma_{in}$)}!@ 
initApp(F,T,BR) :- apply(EP,T,BR),hasEff(EP,F). !@\label{lst:lpIndep:epNotInitiatedStart}!@
kNotInit(F,T,T1,BR) :- not initApp(F,T,BR),								uBr(T1,BR), s(T), fluent(F). !@\label{lst:lpIndep:epNotInitiatedEnd}!@
kNotInit(F,T,T1,BR) :- apply(EP,T,BR), hasPC(EP,F1), 		hasEff(EP,F) ,knows($\neg$F1,T,T1,BR), T1>=T.!@\label{lst:lpIndep:epFalseInitiated}!@
knows(F,T+1,T1,BR) :- knows(F,T,T1,BR), kNotTerm(F,T,T1,BR),T<T1, s(T).!@\label{lst:lpIndep:forwardInertia}!@
knows(F,T-1,T1,BR) :- knows(F,T,T1,BR), 							kNotInit(F,T-1,T1,BR), T>0, T1>=T, s(T).!@\label{lst:lpIndep:backwardInertia}!@
knows(L,T,T1+1,BR) :- knows(L,T,T1,BR),T1<maxS,s(T1).!@\label{lst:lpIndep:inertiaOnKnow}!@
!@$\blacktriangleright$ \f{Sensing and Branching ($\Gamma_{sen}$)}!@
uBr(0,0). uBr(T+1,BR) :- uBr(T,BR), s(T). !@\label{lst:lpIndep:uBr}!@
kw(F,T,T1,BR):- knows(F,T,T1,BR). !@\label{lst:lpIndep:kwStart}!@
kw(F,T,T1,BR):- knows($\neg$F,T,T1,BR).  !@\label{lst:lpIndep:kwEnd}!@
sOcc(T,BR) :- occ(A,T,BR), hasKP(A,_). !@\label{lst:lpIndep:sOcc}!@
leq(BR,BR1) :- BR <= BR1, br(BR), br(BR1).  !@\label{lst:lpIndep:leq}!@
1{nextBr(T,BR,BR1): leq(BR,BR1)}1 :- sOcc(T,BR). !@\label{lst:lpIndep:branching}!@
:- 2{nextBr(T,BR,BR1) :br(BR):s(T)},br(BR1). !@\label{lst:lpIndep:notSameChildBr}!@
uBr(T+1,BR) :- sRes($\neg$F,T,BR). !@\label{lst:lpIndep:senseMakeBranchValid}!@
sRes(F,T,BR) :- occ(A,T,BR),hasKP(A,F),not knows($\neg$F,T,T,BR).!@\label{lst:lpIndep:posSensRes}!@
sRes($\neg$F,T,BR1) :- occ(A,T,BR),hasKP(A,F),not kw(F,T,T,BR), nextBr(T,BR,BR1). !@\label{lst:lpIndep:negSensRes}!@
knows(L,T,T+1,BR) :- sRes(L,T,BR). !@\label{lst:lpIndep:senseAssignValue}!@
knows(F1,T,T1,BR1) :- sOcc(T1,BR), nextBr(T1,BR,BR1), 		knows(F1,T,T1,BR), T1>=T. !@\label{lst:lpIndep:sensingInheritKnowledge}!@
apply(EP,T,BR1) :- sOcc(T1,BR), nextBr(T1,BR,BR1), 				uBr(T1,BR), apply(EP,T,BR), T1>=T.!@\label{lst:lpIndep:sensingInheritEPApp}!@
:-2{occ(A,T,BR):hasKP(A,_)}, br(BR), s(T).!@\label{lst:lpIndep:sensingNoConcurrency}!@
!@$\blacktriangleright$ \f{Plan verification ($\Gamma_{\mathit verify}$)}!@
allWGsAchieved :- uBr(maxS,BR), wGoal(maxS,BR). !@\label{lst:lpIndep:wg1}!@
notAllSGAchieved :- uBr(maxS,BR), not sGoal(maxS,BR). !@\label{lst:lpIndep:sg}!@
planFound :- allWGsAchieved, not notAllSGAchieved.  !@\label{lst:lpIndep:planFound}!@
:- not planFound. !@\label{lst:lpIndep:planFoundNotSM}!@
notGoal(T,BR) :- not wGoal(T,BR), uBr(T,BR). !@\label{lst:lpIndep:gNotAchievedStart}!@
notGoal(T,BR) :- not sGoal(T,BR), uBr(T,BR). !@\label{lst:lpIndep:gNotAchievedEnd}!@
!@$\blacktriangleright$ \f{Plan generation and optimization ($\Gamma_{plan}$)}!@
1{occ(A,T,BR):a(A)}1 :- uBr(T,BR), notGoal(T,BR), 					br(BR), ss(T). % Sequential planning !@\label{lst:lpIndep:seqPlanning}!@
%1{occ(A,T,BR):a(A)} :- uBr(T,BR), notGoal(T,BR), 					br(BR), ss(T). % Concurrent planning !@\label{lst:lpIndep:concPlanning}!@
#minimize {occ(_,_,_) @ 1} 	 % Optimal planning !@\label{lst:lpIndep:optimizePlans}!@
\end{lstlisting}

(2) a fluent is known not to be initiated if an effect proposition with that fluent is applied, but one of its conditions is known not to hold (\li{lst:lpIndep:epFalseInitiated}).
Note that this requires the concurrency restriction (\li{lst:lpIndep:noSimilarConcurrent}).
Having defined \co{kNotInit/4} and \co{kNotTerm/4} we can formulate forward inertia (\li{lst:lpIndep:forwardInertia}) and backward inertia (\li{lst:lpIndep:backwardInertia}). 
Two respective rules for inertia of false fluents are not listed for brevity.
We formulate \i{forward propagation} of knowledge in \li{lst:lpIndep:inertiaOnKnow}.
That is, if at $t'$ it is known that $f$ was true at $t$, then this is also known at $t'+1$. 

\noindent  \f{F3.\quad  Sensing and Branching  ($\Gamma_{sen}$)}\quad If sensing occurs, then each possible outcome of the sensing uses one branch.
\co{uBr($t$,$br$)} denotes that branch $br$ is used at step $t$. 
Predicate \co{kw/4} in \lir{lst:lpIndep:kwStart}{lst:lpIndep:kwEnd} is an abbreviation for \i{knowing whether}. 
We use \co{sOcc($t$,$br$)} to state that a sensing action occurred at $t$ in $br$ (\li{lst:lpIndep:sOcc}). 
By \co{leq($br$,$br'$)} the partial order of branches is precomputed (\li{lst:lpIndep:leq}); it is used in the choice rule \li{lst:lpIndep:branching} to ``pick'' a valid child branch when sensing occurs. Two sensing actions are not allowed to pick the same child branch (\li{lst:lpIndep:notSameChildBr}).
%
Lines \ref{lst:lpIndep:posSensRes}-\ref{lst:lpIndep:negSensRes} assign the positive sensing result to the current branch and the negative result to the child branch. Sensing results affect knowledge through \li{lst:lpIndep:senseAssignValue}.  Line \ref{lst:lpIndep:sensingInheritKnowledge} represents inheritance:
Knowledge and application of EPs is transferred from the original branch to the child branch (\li{lst:lpIndep:sensingInheritEPApp}).
%
Finally, in line \li{lst:lpIndep:sensingNoConcurrency}, we make the restriction that two sensing actions cannot occur concurrently.


\noindent \f{F4.\quad Plan Verification ($\Gamma_{\mathit verify}$)}\quad
\Lir{lst:lpIndep:wg1}{lst:lpIndep:planFoundNotSM} handle that weak goals must be achieved in only one branch and strong goals in all branches.
Information about nodes where goals are not yet achieved (\lir{lst:lpIndep:gNotAchievedStart}{lst:lpIndep:gNotAchievedEnd}) is used in the plan generation part for pruning.


\noindent \f{F5.\quad Plan Generation and Optimization ($\Gamma_{plan}$)}\quad
\Li{lst:lpIndep:seqPlanning} and \li{lst:lpIndep:concPlanning} implement sequential and concurrent planning respectively. Optimal plans in terms of the number of actions are generated with the optimization statement \li{lst:lpIndep:optimizePlans}.
\vspace{-12pt}
\subsection{Plan Extraction from Stable Models}\label{ssec:PlanExtraction}
\vspace{-7pt}
A conditional plan is determined by a set of \co{occ/3}, \co{nextBr/3} and \co{sRes/3} atoms.
\vspace{-5pt}
\begin{Definition}[{\bf\small Planning as ASP Solving}]
{\small Let $S$ be a stable model 
for the logic program LP(\mc{P}), then $p$ solves the planning problem \mc{P} if $p$ is exactly the subset containing all \co{occ/3}, \co{nextBr/3} and \co{sRes/3} atoms of $S$.}
\end{Definition}
\vspace{-5pt}
\noindent For example, consider the atoms \co{occ($a_0$,$t$,$br$}), \co{sRes($f$,$t$,$br$)}, \co{sRes($\neg f$, $t$,$br'$)}, \co{nextBr($t$,$br$,$br'$)}, \co{occ($a_1$,$t+1$,$br$}) and \co{occ($a_2$,$t+1$,$br'$}). With a syntax as in \citep{Tu2007}, this is equivalent to the conditional plan \co{$a_0$;[if $f$ then $a_1$ else $a_2$]}. 

\vspace{-15pt}
\subsection{Complexity of h-approximation}
\label{ssec:complexity}
\vspace{-7pt}
According to \citep{Tu2007}, we investigate the complexity for a limited number of sensing actions, and feasible plans. That is, plans with a length that is polynomial \wrt the size of the input problem.
\begin{Theorem}[{\bf\small (Optimal) Plan Existence}]
\label{theo:approxPlanExistsIsNP}\small
The plan existence problem for the h-approximation is in NP and finding an optimal plan is in $\Delta_2^P$.
\end{Theorem}
\vspace{-5pt}
\noindent \f{Proof Sketch:}
\vspace{-5pt}
{\small
The result emerges directly from the complexity properties of ASP (\eg \citep{Gebser2012}).
\enum{
\item The translation of an input problem via (\ref{lst:lp-declaration}-\ref{lst:lp-goals}) is polynomial.
\item Grounding the normal logic program is polynomial because the arity of predicates is fixed and \co{maxS} and \co{maxBr} are bounded due the polynomial plan size. 
\item Determining whether there exists a stable model for a normal logic program is NP-complete.
\item Finding an optimal stable model for a normal logic program is $\Delta_2^P$-complete.
}
}

\subsection{Translation Optimizations}\label{sec:transOpt}
Although optimization of \hpx\ is not in the focus at this stage of our work we want to note two obvious aspects:
(1) By avoiding \emph{unnecessary action execution}, e.g.~opening a door if it is already known to be open, search space is pruned significantly.
(2) Some domain specificities (e.g., connectivity of rooms) are considered as \emph{static relations}. For these, we modify translation rules (\ref{lst:lp-actionExcondition}) 
(executability conditions) 
and (\ref{lst:lp-initKnowledge}) 
(value propositions), 
such that \co{knows/4} is replaced by \co{holds/1}.

\section{A Temporal Query Semantics for \Ak}
\label{sec:relToPWS}
%
\hpx\ is not just an approximation to \pws\ as implemented in \Ak. 
It is more expressive in the sense that \hpx\ allows for propositions about the past, \eg 
``at step 5 it is known that the door was open at step 3''.
To find a notion of soundness of \hpx\ with \Ak\ (and hence \pws-based approaches in general), we define a \i{temporal query semantics} (\Ak$^{TQS}$) that allows for reasoning about the past.
%
The syntactical mapping between \Ak\ and \hpx\ is presented in the following table:
%

\noindent \begin{small}
	\begin{tabular}{p{.15\textwidth}|p{.33\textwidth}|p{.48\textwidth}}
						& \Ak									& \hpx\ PDDL dialect\\ \hline
		Value prop.		& \f{initially}($l^{init})$ 						& \cos{(:init $l^{init}$)} \\ \hline
		Effect prop.		& \f{causes}($a$,$l^e$, \{$l^c_1 \hdots l^c_n$\})	& \cos{(:action $a$} \cos{:effect when (and $l^c_1 \hdots l^c_n$) $l^e$)} \\ \hline
		Executability	& \f{executable}($a$, \{$l^{ex}_1,\hdots,l^{ex}_n$\})	& \cos{(:action $a$} \cos{:executable (and$\,\,l^{ex}_1\hdots l^{ex}_n$))} \\ \hline
		Sensing	& \f{determines} ($a$,\{$f$,$\neg f$\})			& \cos{(:action $a$} \cos{:observe $f$)} \\ \hline
	\end{tabular}
	\end{small}
%

\smallskip
\noindent An \Ak\ domain description D can always be mapped to a corresponding \hpx\ domain specification due to the syntactical similarity. 
Note that for brevity we do not consider executability conditions in this section.
Their implementation and intention is very similar in h-approximation and \Ak.
Further we restrict the \Ak\ semantics to allow to sense the value of only one single fluent with one action. 

\subsection{Original \Ak\ Semantics by \citet{Son2001}} 
\Ak\ is based on a transition function which maps an action and a so-called c-state to a c-state. 
A c-state $\delta$ is a tuple $\rab{u,\Sigma}$, where $u$ is a state (a set of fluents) and $\Sigma$ is a k-state (a set of possible belief states).
If a fluent is contained in a state, then its value is $true$, and $false$ otherwise.
Informally, $u$ represents how the world is and $\Sigma$ represents the agent's belief.
In this work we assume grounded c-states for \Ak, \ie $\delta = \rab{u,\Sigma}$ is grounded if $u \in \Sigma$. 
The transition function for non-sensing actions and without considering executability is:
{\small
\begin{align}
&\Phi(a,\rab{u,\Sigma}) = \rab{Res(a,u), \{Res(a,s') | s' \in \Sigma \}} \text{ where } \label{eq:akTransFunc}\\[-3pt]
&~~~~~Res(a,s) = s \cup E^+_a(s) \setminus E^-_a(s) \text{ where } \label{eq:akResFunc}\\[-3pt]
&~~~~~~~~E^+_a(s) = \{ f| \text{ $f$ is the effect literal of an EP and all condition literals hold in $s$}\} \notag \\[-3pt]
&~~~~~~~~E^-_a(s) = \{ \neg f| \text{ $\neg f$ is the effect literal of an EP and all condition literals hold in $s$}\} 
\end{align}
}%
$Res$ reflects that if all conditions of an effect proposition hold, then the effect holds in the result.
The transition function for sensing actions is:
{\small
\begin{align}
&\Phi(a,\rab{u,\Sigma}) = \rab{u, \{s | (s \in \Sigma) \wedge (f \in s \Leftrightarrow f \in u) \}} \label{eq:akTransFuncSensing}
\end{align}
}
For convenience we introduce the following notation for a k-state $\Sigma$:
{\small
\begin{equation}
\label{eq:Ak-simplifySigmaEntailment}
\Sigma \models f \text{ iff } \forall s \in \Sigma: f \in s \text{ and }
\Sigma \models \neg f \text{ iff } \forall s \in \Sigma: f \cap s = \emptyset
\end{equation}
}
It reflects that a fluent is known to hold if it holds in all possible worlds $s$ in $\Sigma$.

\subsection{Temporal Query Semantics -- \Ak$^{TQS}$}
Our approach is based on a re-evaluation step with a similar intuition as the \emph{update operator} ``$\circ$'' in \citep{Vlaeminck2012}:
Let $\Sigma_0 = \{s^0_0,\hdots,s^{|\Sigma_0|}_0\}$ be the set of all possible initial states of a (complete) initial c-state of an \Ak\ domain D.
Whenever sensing happens, the transition function will remove some states from the k-state, \ie\
$\Phi([a_1;\hdots;a_n], \delta_0) = \rab{u_n,\Sigma_n}$, where $\Sigma_n = \{s^0_n,\hdots,s^{|\Sigma_n|}_n\}$ and $|\Sigma_0| \geq |\Sigma_n|$. 
To reason about the past, we re-evaluate the transition.
Here, we do not consider the complete initial state, but only the subset $\Sigma^n_0$ of initial states which ``survived'' the transition of a sequence of actions.
If a fluent holds in all states of a k-state $\Sigma^n_t$, where $\Sigma^n_t$ is the result of applying $t \leq n$ actions on $\Sigma^n_0$, then after the $n$-th action, it is known that a fluent holds after the $t$-th action.
%
\begin{Definition}
\begin{sloppypar}
{\small
Let $\alpha=[a_1;\hdots;a_n]$ be a sequence of actions and $\delta_0$ be a possible initial state, such that 
\noindent $\Phi([a_1;\hdots;a_n], \delta_0) = \delta_n = \rab{u_n, \Sigma_n}$.
We define $\Sigma^n_0$ as the set of initial belief states in $\Sigma_0$ which are valid after applying $\alpha$:
$\Sigma^n_0 = \{s_0 | s_0 \in \Sigma_0 \wedge Res(a_n, Res(a_{n-1}, \hdots, Res(a_1, s_0) \hdots )) \in \Sigma_n\}$.%
\footnote{Consider that according to (\ref{eq:akResFunc}) $Res(a,s) = s$ if $a$ is a sensing action.}
We say that
\begin{center}
\emph{$\rab{l,t}$ is known to hold after $\alpha$ on $\delta_0$} 
\end{center}
if 
$\Sigma^n_t \models l$ where $\rab{u_t,\Sigma^n_t} = \Phi([a_1;\hdots;a_t], \rab{u_0,\Sigma^n_0}) \text{ and } t \leq n$
}
\end{sloppypar}
\end{Definition}


\subsection{Soundness \wrt \Ak$^{TQS}$}
The following conjecture considers soundness for the projection problem for a sequence of actions:%

%
%
%
\begin{Conjecture}\label{prop:tqsSoundnessSimple}
{\small
\begin{sloppypar}
Let $D$ be a domain specification and $\alpha = [a_1;\hdots;a_n]$ be a sequence of actions. 
Let $LP(D) = [\Gamma_{in} \cup\Gamma_{sen}\cup\Gamma_{conc}\cup \Gamma_{ini}\cup\Gamma_{act}]$ be a \hpx-logic program without rules for plan generation ($\Gamma_{plan}$), plan verification ($\Gamma_{\mathit verify}$) and goal specification ($\Gamma_{goal}$). 
Let $\Gamma_{occ}^n$ contain rules about action occurrence in valid branches, \ie  $\Gamma_{occ}^n = \{occ(a_0,0,BR) \lam uBr(0,BR). , \hdots, occ(a_{n},n,BR) \lam uBr(n,BR).\}$
Then for all fluents $f$ and all steps $t$ with $0 \leq t \leq n$, there exists a branch $br$ such that: 
\begin{equation}
\label{eq:soundnessSimpleHypo}
\setlength\abovedisplayskip{2pt}
\setlength\belowdisplayskip{2pt}
\begin{aligned}
\text{ if } \co{knows($l$,$t$,$n$,$br$)} \in SM[LP(D) \cup \Gamma_{occ}^n] 
\text{ then } \Sigma^n_t \models l
~~~\text{  with } t \leq n.
\end{aligned}
\end{equation}
where $SM[LP(D) \cup \Gamma_{occ}^n]$ denotes the stable model of the logic program.%
\end{sloppypar}
}
\end{Conjecture}
%
%

The following observation is essential to formally investigate soundness:
\vspace{-6pt}
\begin{Observation}\label{obs:ReasonsForKnowledge}
{\small
We investigate $\Gamma_{hpx}$ (Listing \ref{lst:lpIndep}) and $\Gamma_{world}$ and observe that an atom \\
\co{knows($f$,$t$,$n$,$br$)} can only be produced by 
\ienum{
\item \label{enum:obs:initKnow} Initial Knowledge (\ref{lst:lp-initKnowledge})
\item \label{enum:obs:sensing} Sensing (\li{lst:lpIndep:senseAssignValue})
\item \label{enum:obs:inheritance} Inheritance (\li{lst:lpIndep:sensingInheritKnowledge})
\item \label{enum:obs:fwdInit} Forward inertia (\li{lst:lpIndep:forwardInertia})
\item \label{enum:obs:bckInit} Backward inertia (\li{lst:lpIndep:backwardInertia})
\item \label{enum:obs:fwdProp} Forward propagation (\li{lst:lpIndep:inertiaOnKnow})
\item \label{enum:obs:causality} Causation (\ref{lst:lp-effCondTrans1})
\item \label{enum:obs:posPost} Positive postdiction (\ref{lst:lp-effCondTrans2})
or
\item \label{enum:obs:negPost} Negative postdiction (\ref{lst:lp-effCondTrans3}).
}
}
\end{Observation}

\begin{sloppypar}
\subsubsection{Conditional Proof Sketch}
{
This conditional proof sketch contains circular dependencies and hence can not be considered as a full proof. 
However, it does provide evidence concerning the correctness of Conjecture \ref{prop:tqsSoundnessSimple}.

\noindent To demonstrate soundness we would investigate each item \i{(\ref{enum:obs:initKnow}--\ref{enum:obs:negPost})} in Observation \ref{obs:ReasonsForKnowledge} and show that if
$\co{knows($f$,$t$,$n$,$br$)} \in SM[LP(D) \cup \Gamma_{occ}^n] $ produced by this item, then $\Sigma^n_t \models f$ must hold for some $br$. 
However, for reasons of brevity we consider only some examples \i{(%
\ref{enum:obs:sensing}, \ref{enum:obs:bckInit}, \ref{enum:obs:posPost})} for positive literals $f$ and without executability conditions:
\vspace{-5pt}
\enum{
\item Sensing \i{(\ref{enum:obs:sensing})}. \quad The soundness proof for sensing is by induction over the number of sensing actions. 
For the base step we have that $br=0$ (\li{lst:lpIndep:uBr}).
	A case distinction for positive ($f \in u$) and negative ($f \not\in u$) sensing results is required: 
	With (\lir{lst:lpIndep:posSensRes}{lst:lpIndep:negSensRes}) the positive sensing result is applied to the original branch $br$ and the negative result is applied to a child branch determined by \co{nextBr/3}. 
	The hypothesis holds \wrt one of these branches. 
	The \Ak\ restriction that sensing and non-sensing actions are disjoint ensures that the sensed fluent did not change during the sensing. 
	Hence, its value after sensing must be the same as at the time it was sensed. 
	This coincides with our semantics where sensing returns the value of a fluent at the time it is sensed.  


%
\item Backward Inertia \i{(\ref{enum:obs:bckInit})}.\quad 
Backward inertia (\li{lst:lpIndep:backwardInertia}) generates $\co{knows($f$,$t$,$n$,$br$)}$ with 
$t < n$ if both of the following is true:
					\itz{
						\item[A:] $\co{knows($f$,$t+1$,$n$,$br$)}$ is an atom in the stable model. 
						If this is true and we assume that the conjecture holds for $t+1$, then $\Sigma^{n}_{t+1} \models f$.
						\item[B:] $\co{kNotInit($f$,$t$,$n$,$br$)}$ is an atom in the stable model. 
						This again is only true if \i{(i)} no action with an EP with the effect literal $f$ is applied at $t$ (\lir{lst:lpIndep:epNotInitiatedStart}{lst:lpIndep:epNotInitiatedEnd}) or 
							\i{(ii)} an action with an EP with the effect literal $f$ is applied at $t$, but this EP has at least one condition literal which is known not to hold (\li{lst:lpIndep:epFalseInitiated}). 				
							As of the result function (\ref{eq:akResFunc}) this produces in both cases that $\forall s_t^n \in \Sigma^n_t : E^+_a(s_t^n) = \emptyset$. 
					}
					With A: $\Sigma^{n}_{t+1} \models f$ and  B: $\forall s_t^n \in \Sigma^n_t : E^+_a(s_t^n) = \emptyset$, we can tell by the transition function (\ref{eq:akTransFunc}) that $\Sigma^{n}_{t} \models f$ and the case of backward inertia is conditionally proven if the conjecture holds for $\co{knows($f$,$t+1$,$n$,$br$)}$. 
\item Positive Postdiction \i{(\ref{enum:obs:posPost})}. \quad 
Positive postdiction (\ref{lst:lp-effCondTrans2}) generates an atom $\co{knows($f^c_i$,$t$,$n$,$br$)}$ if  \co{apply($ep$,$t$,$br$)}, \co{knows($f^e$,$t+1$,$n$,$br$)} and \co{knows($\ol{f^e}$,$t$,$n$,$br$)} with $t < n$ and where $f^c_i$ is a condition literal and $f^e$ is an effect literal of $ep$. 
We can show that positive postdiction generates correct results for the condition literals if Conjecture \ref{prop:tqsSoundnessSimple} holds for knowledge about the effect literal: 
That is, if we assume that \i{(i)} $\Sigma^n_{t+1} \models f^e$ and \i{(ii)} $\Sigma^n_t \models \ol{f^e}$, then with the result function (\ref{eq:akResFunc}), \i{(i)} and \i{(ii)} can only be true if $E^+_a(s^n_t) = f^e$ for all $s^n_t \in \Sigma^n_t$. 
Considering the restriction that only one EP with a certain effect literal $f^e$ may be applied at once (\li{lst:lpIndep:noSimilarConcurrent}), $E^+_a(s^n_t) = f^e$ can only hold if for all conditions $f^c_i$: $\Sigma^n_t \models f^c_i$.

				}
\vspace{-5pt}
The case for causation, negative postdiction, forward inertia, etc.~is similar.
}
\end{sloppypar}


\section{Evaluation and Case-Study}
\label{sec:evalAndCasStudy}
In order to evaluate practicability of \hpx\ we compare our implementation with the ASCP planner by \citet{Tu2007} and show an integration of our planning system in a smart home assistance system.

\f{Comparison with ASCP.}\quad
We implemented three well known benchmark problems for \hpx\ and the 0-approximation based ASCP planner:%
\footnote{We used an Intel i5 (2GHz, 6Gb RAM)  machine running \i{clingo} \citep{Gebser2012} with Windows 7. Tests were performed for a fixed plan length and width.}
\i{Bomb in the toilet} 
(\eg \citep{Hoffmann2005}; $n$ potential bombs need to be disarmed in a toilet), 
\i{Rings} 
(\eg \citep{Cimatti2003}; in $n$ ringlike connected rooms windows need to be closed/locked), and
\i{Sickness} 
(\eg \citep{Tu2007}; one of $n$ diseases need identified with a paper color test).
While \hpx\ outperforms ASCP for the Rings problem (\eg $\approx$ 10s to 170s for 3 rooms), ASCP outperforms \hpx\ for the other domains (\eg $\approx$ 280s to 140s for 8 bombs and $\approx$ 160s to 1360s for 8 diseases).
For the first problem, \hpx\ benefits from static relations and for the latter two problems ASCP benefits from a simpler knowledge representation and the ability to sense the paper's color with a single action where \hpx\ needs $n-1$ actions.
In both ASCP and \hpx\ grounding was very fast and the bottleneck was the actual solving of the problems.

\begin{figure*}[t!]
\centering
\begin{minipage}[t]{0.325\textwidth}%
\includegraphics[width=\textwidth]{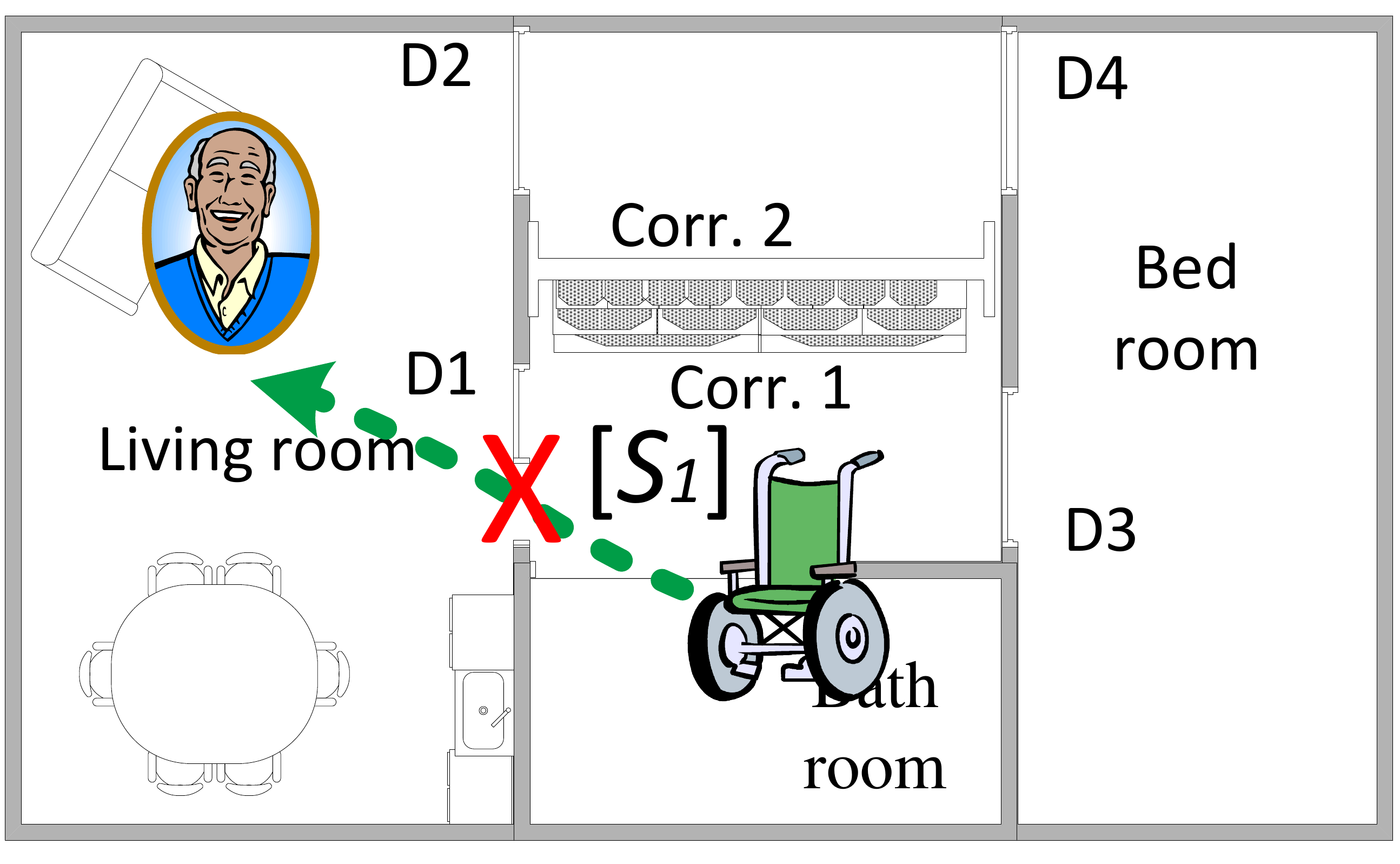}
\small
\end{minipage}%
\hspace{.002\textwidth}
\begin{minipage}[t]{0.325\textwidth}%
\includegraphics[width=\textwidth]{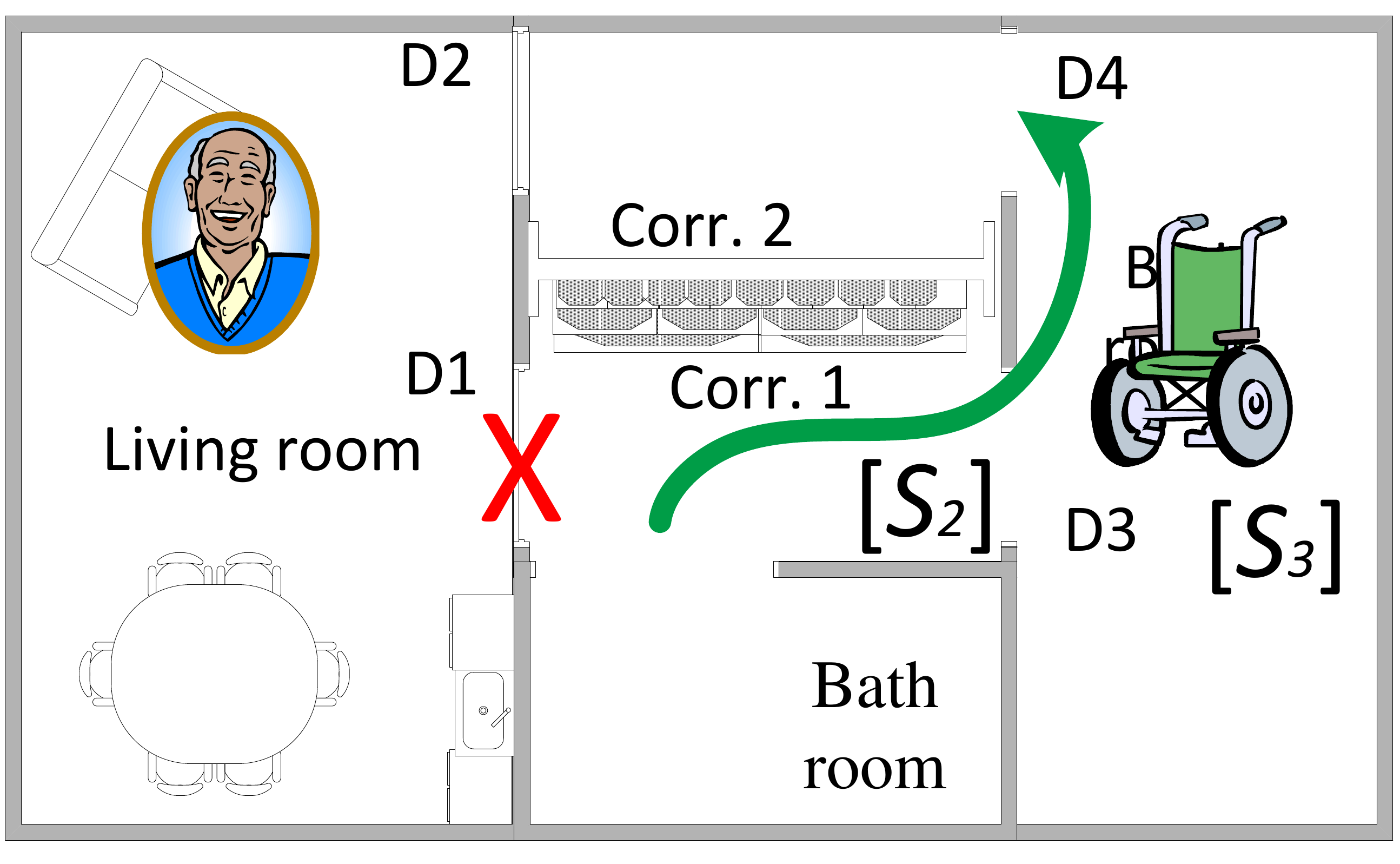}
\small
\end{minipage}%
\hspace{.002\textwidth}
\begin{minipage}[t]{0.325\textwidth}%
\includegraphics[width=\textwidth]{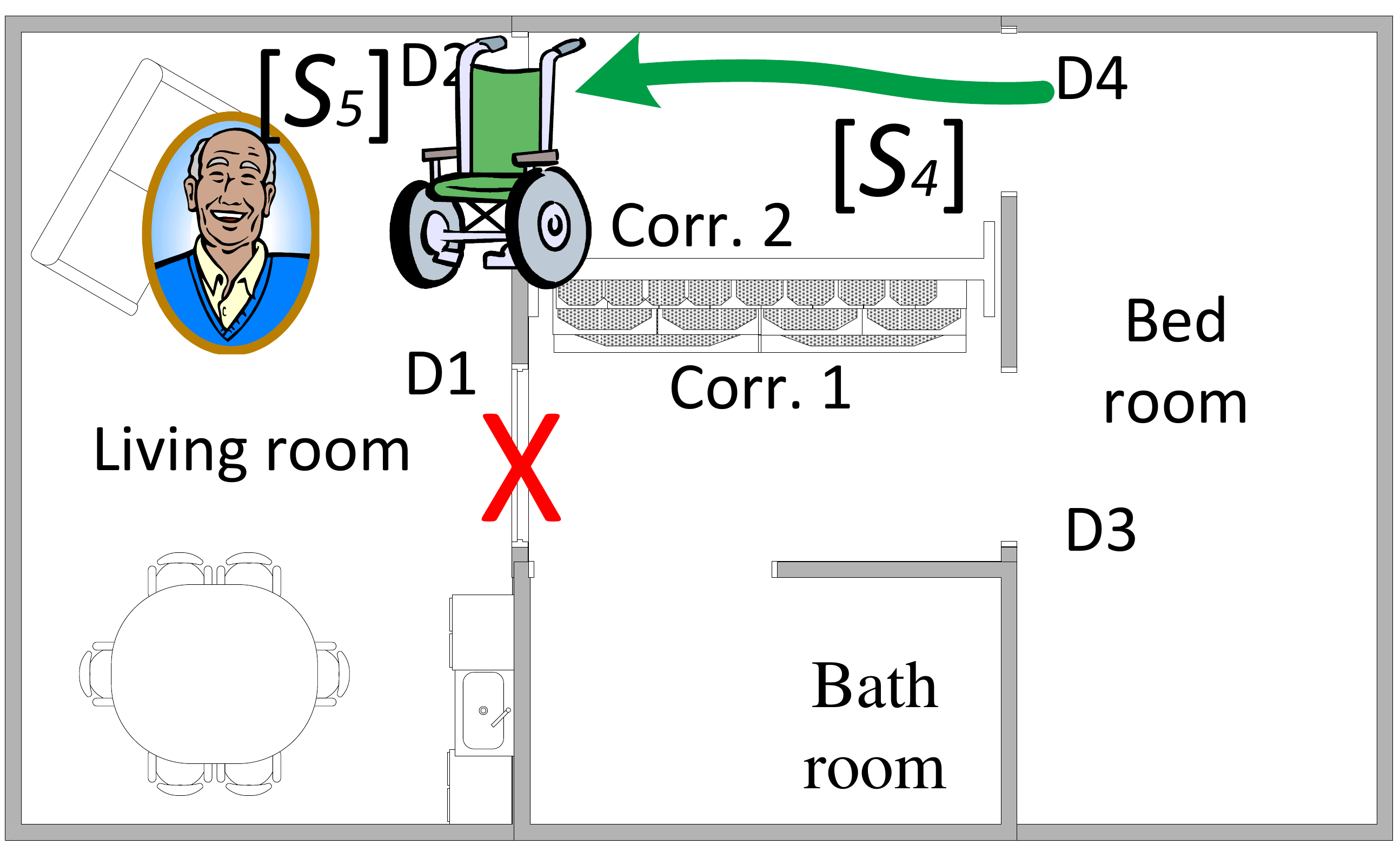}
\small
\end{minipage}%

\begin{minipage}[t]{0.1588\textwidth}%
\includegraphics[width=\textwidth]{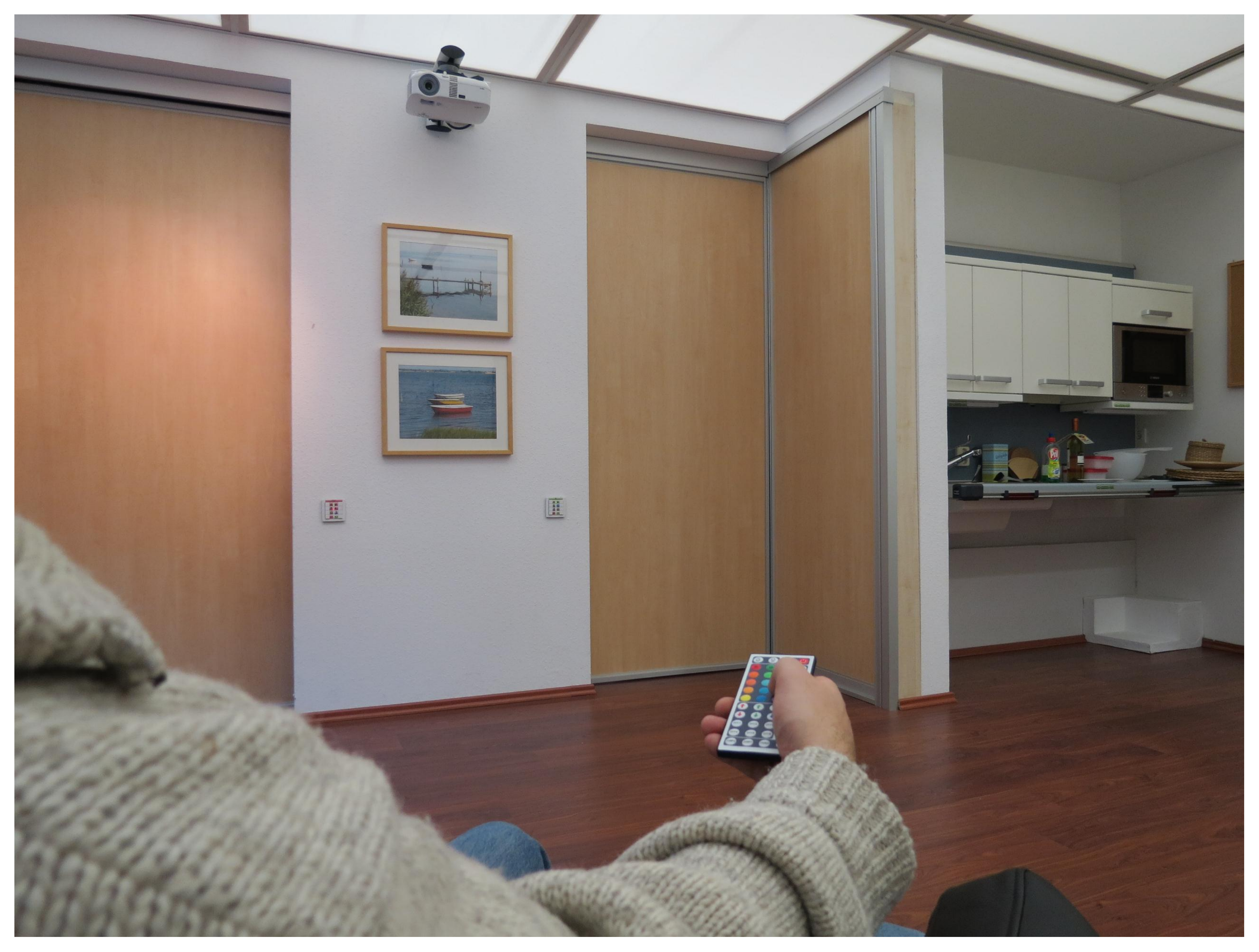}
\small
[$S_0$]
\end{minipage}%
\hspace{.001\textwidth}
\begin{minipage}[t]{0.1588\textwidth}%
\includegraphics[width=\textwidth]{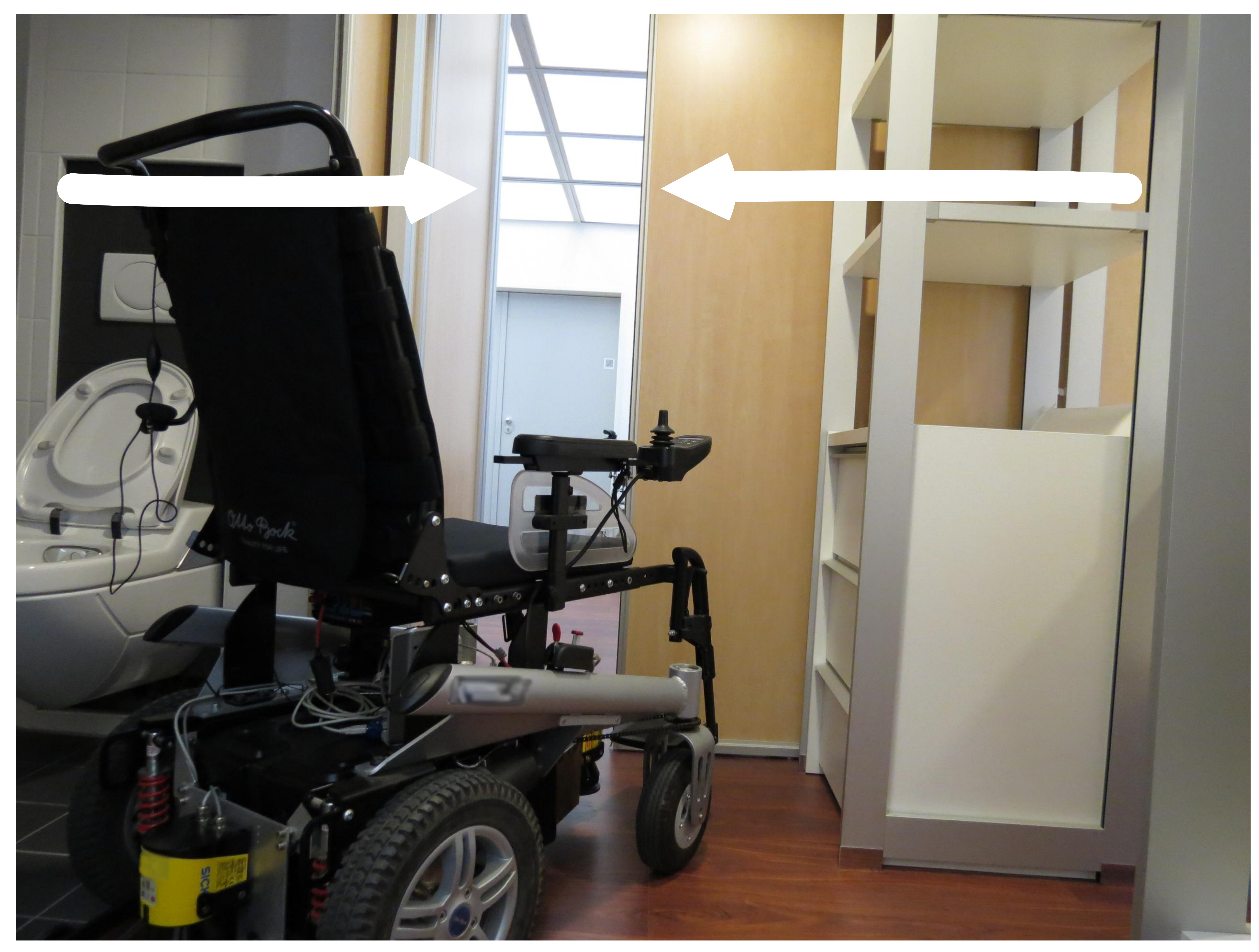}
\small
[$S_1$]
\end{minipage}%
\hspace{.001\textwidth}
\begin{minipage}[t]{0.1588\textwidth}%
\includegraphics[width=\textwidth]{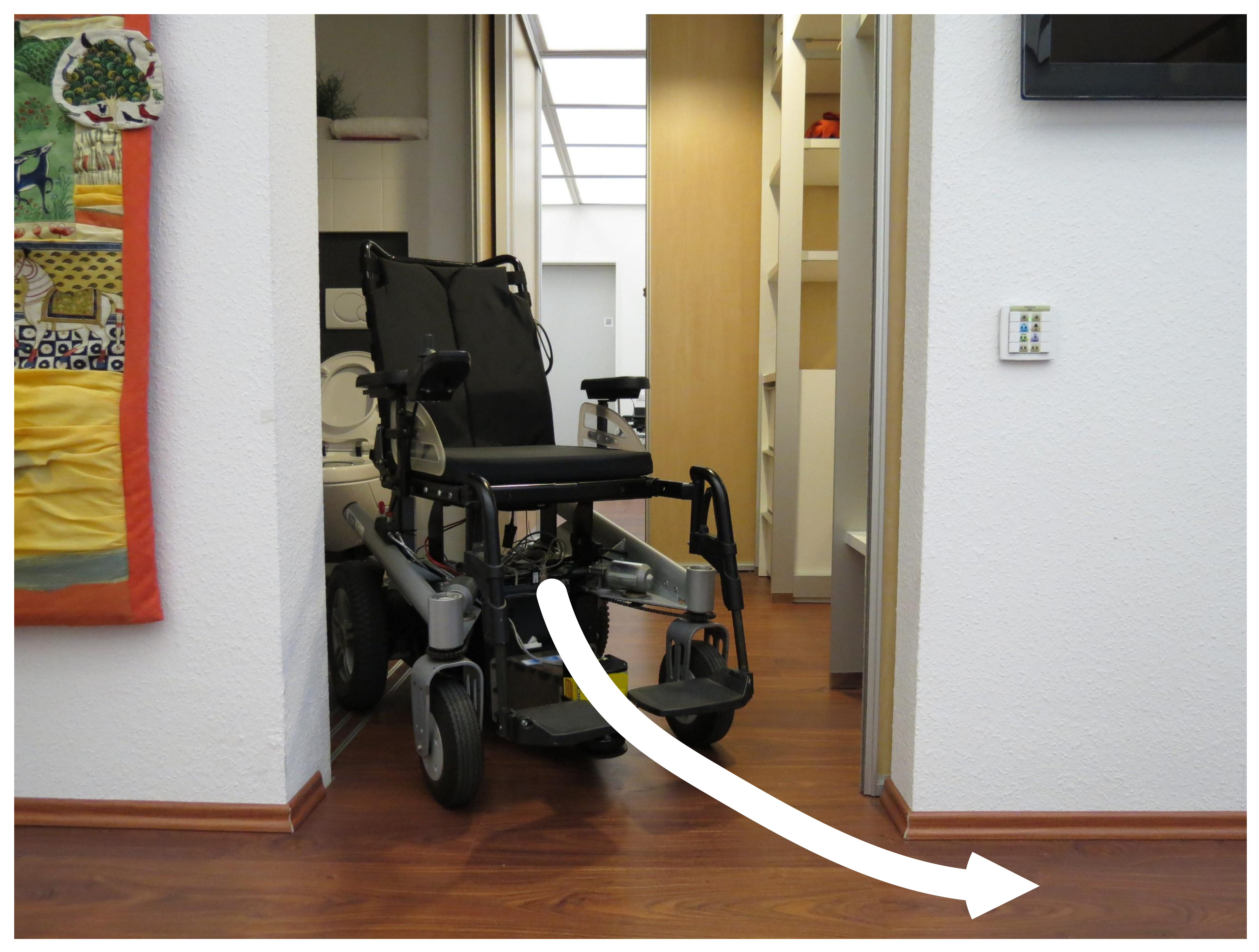}
\small
[$S_2$]
\end{minipage}%
\hspace{.001\textwidth}
\begin{minipage}[t]{0.1588\textwidth}%
\includegraphics[width=\textwidth]{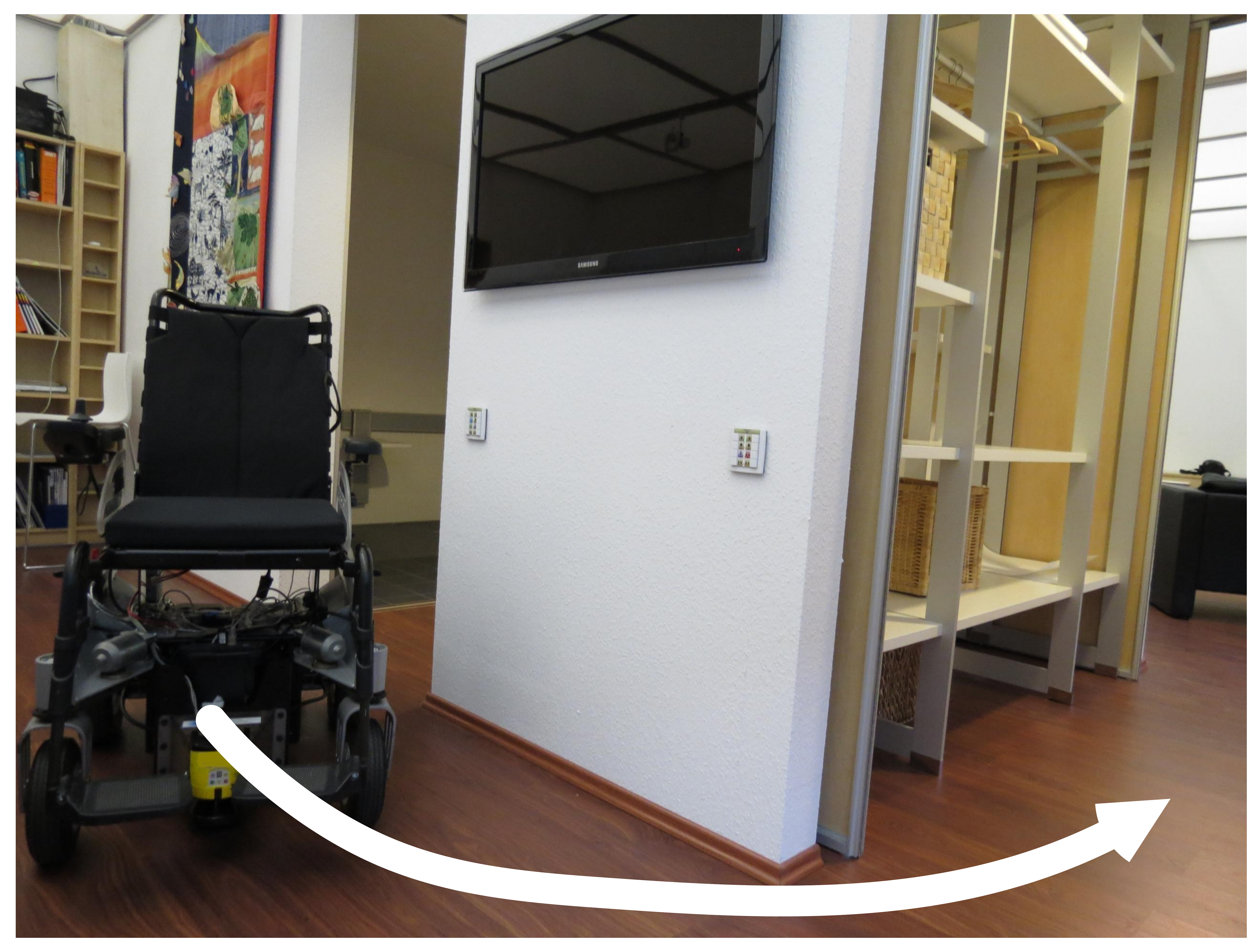}
\small
[$S_3$]
\end{minipage}%
\hspace{.001\textwidth}
\begin{minipage}[t]{0.1588\textwidth}%
\includegraphics[width=\textwidth]{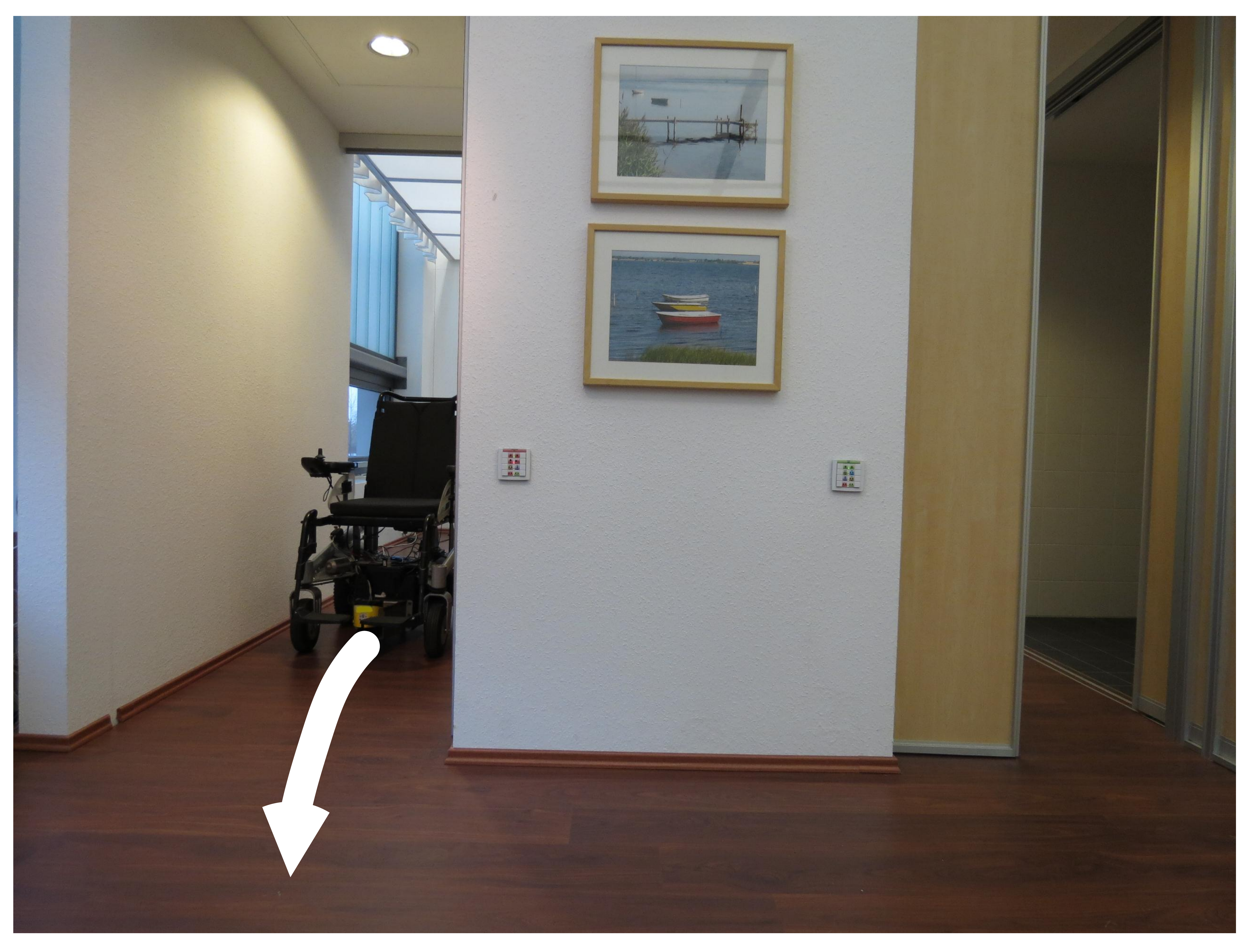}
\small
[$S_4$]
\end{minipage}%
\hspace{.001\textwidth}
\begin{minipage}[t]{0.1588\textwidth}%
\includegraphics[width=\textwidth]{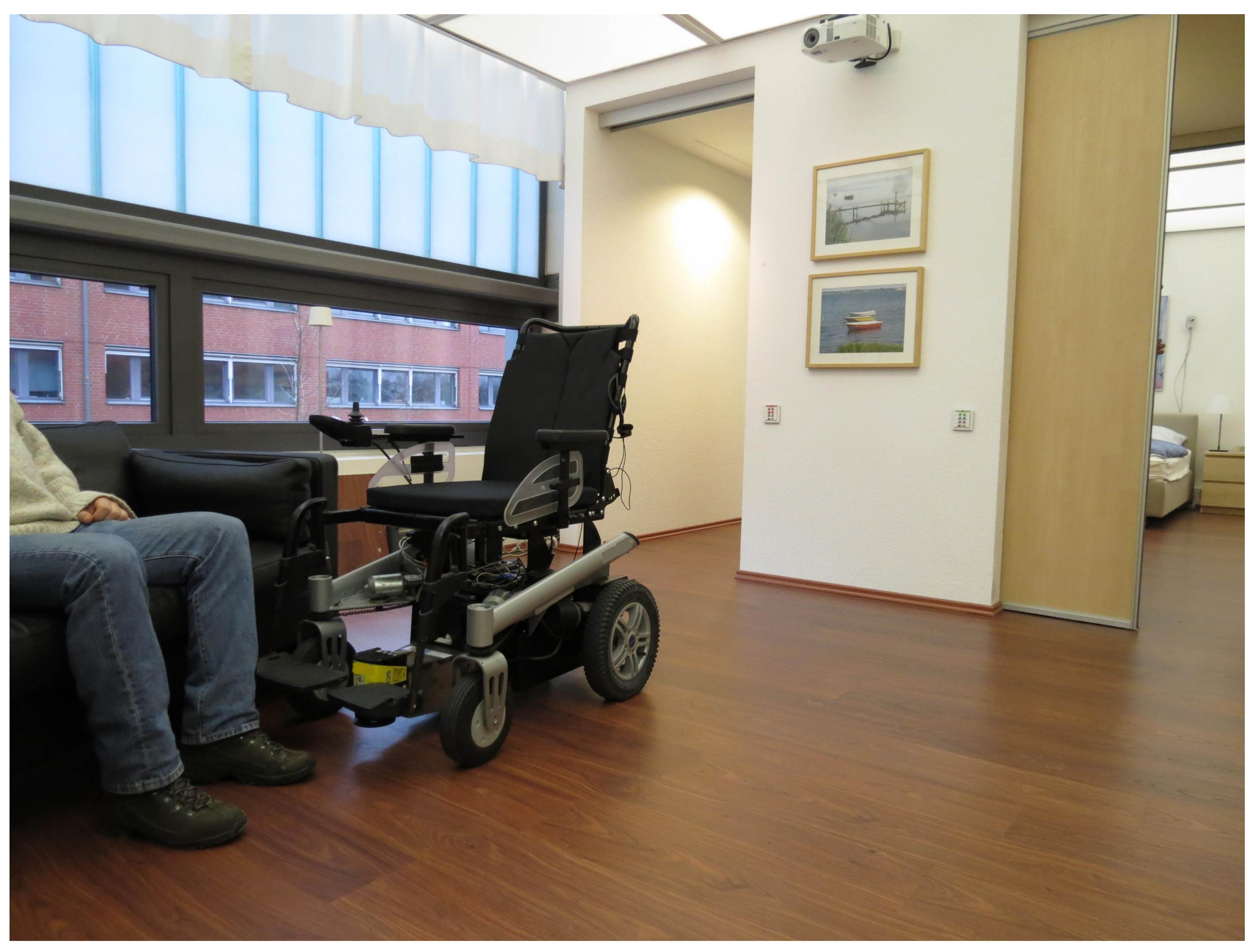}
\small
[$S_5$]
\end{minipage}%
\smallskip
\caption{The wheelchair operating in the smart home BAALL}%
\label{fig:useCasePhotos}%
\end{figure*}

\f{Application in a Smart Home.}\quad
The \hpx\ planning system has been integrated within a larger software framework for smart home control in the Bremen Ambient Assisted Living Lab (BAALL) \citep{Krieg-Bruckner2010}. 
We present a use-case involving action planning in the presence of abnormalities for an robotic wheelchair: The smart home has (automatic) sliding doors, and sometimes a box or a chair accidentally blocks the door such that it opens only half way. In this case, the planning framework should be able to postdict such an abnormality and to follow an alternative route. The scenario is illustrated in Fig. \ref{fig:useCasePhotos}. 

\begin{figure*}[t!]
		\hspace{-10.5pt}\includegraphics[width=1.04\textwidth]{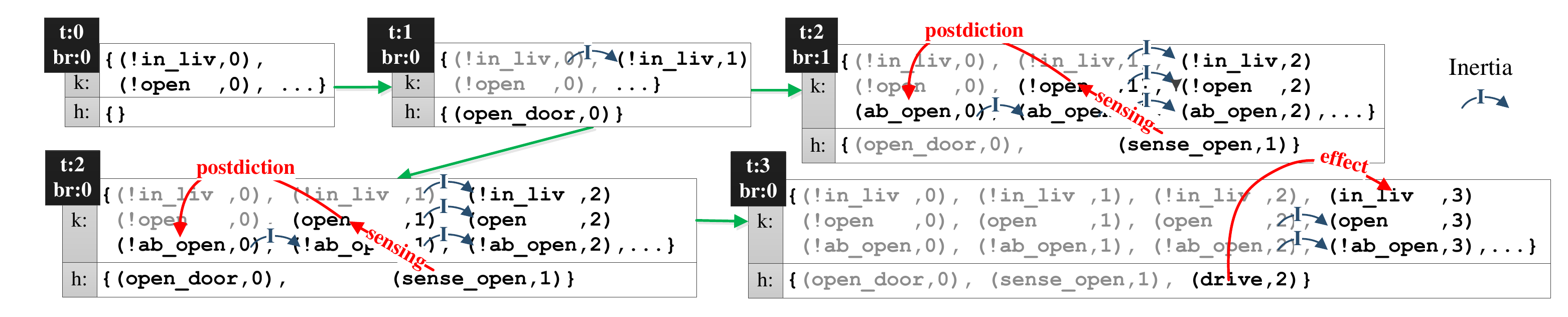}
		\caption{Abnormality detection as postdiction with \emph{h-approximation}}
	\label{fig:postdictionScheme}
\end{figure*}

Consider the situation where a person instructs a command to the wheelchair (e.g., to reach location; [$S_0$]).
An optimal plan to achieve this goal is to pass D1. A more error tolerant plan is: 
{Open D1} and verify if the action succeeded by sensing the door status [$S_1$]; 
\i{If} the door is open, drive through the door and approach the user.
\i{Else} there is an abnormality: 
Open and pass D3 [$S_2$]; drive through the bedroom [$S_3$]; pass D4 and D2 [$S_4$]; and finally approach the sofa [$S_5$].%
\footnote{Abnormalities are considered on the alternative route but skipped here for brevity.}
If it is behind the door then the door was open. 
For this particular use-case, a sub-problem follows:

\vspace{-8pt}
\begin{center}
\begin{lstlisting}[style=pddl,label=lst:pddl,escapechar=@]
(:action open_door :effect when $\neg$ab_open open)@\label{lst:pddl:actOpenDoorStart}@
(:action drive :executable (and open	$\neg$in_liv) 
							 :effect in_liv) @\label{lst:pddl:actDriveEnd}@
(:action sense_open :observe open) @\label{lst:pddl:actSenseOpenStart}@
(:init $\neg$in_liv $\neg$open)      (:goal weak in_liv)@\label{lst:pddl:initGoal}@
\end{lstlisting}
\end{center}
\vspace{-8pt}
\begin{sloppypar}
The solution to this subproblem is depicted in Fig.~\ref{fig:postdictionScheme} (see also state $S_1$  in Fig.~\ref{fig:useCasePhotos}).
There is an autonomous robotic wheelchair outside the living room (\co{$\neg$in\_liv}) and the weak goal is that the robot is inside the living room.
The robot can open the door (\co{open\_door}) to the living room. 
Unfortunately, opening the door does not always work, as the door may be jammed, \ie there may be an abnormality.
However, the robot can perform sensing to verify whether the door is open (\co{sense\_open}).
Figure \ref{fig:postdictionScheme} illustrates our postdiction mechanism. 
Initially (at $t=0$ and $br=0$) it is known that the robot is in the corridor at step $0$. 
The first action is opening the door, \ie the stable model contains the atom \co{occ(open\_door,0,0)}. 
Inertia holds for \co{$\neg$in\_liv}, because nothing happened that could have initiated \co{$\neg$in\_liv}. 
The rules in \lir{lst:lpIndep:epNotInitiatedStart}{lst:lpIndep:epNotInitiatedEnd} trigger \co{kNotInit(in\_liv,0,0,0)} and \li{lst:lpIndep:inertiaOnKnow} triggers \co{knows($\neg$in\_liv,0,1,0)}, such that in turn the forward inertia rule (\li{lst:lpIndep:forwardInertia}) causes atom \co{knows($\neg$in\_liv,1,1,0)} to hold.
Next, sensing happens, \ie \co{occ(sense\_open,1,0)}. 
According to the rule in \li{lst:lpIndep:posSensRes}, the positive result is assigned to the original branch and \co{sRes(open,1,0)} is produced.
According to the rule in \li{lst:lpIndep:negSensRes}, the negative sensing result at step $t$ in branch $br$ is assigned to some child branch $br'$ (denoted by \co{nextBr($t$,$br$,$br'$)}) with $br' > br$ (\li{lst:lpIndep:branching}). In the example we have: \co{sRes($\neg$open,1,1)}, and due to \li{lst:lpIndep:senseAssignValue} we have \co{knows($\neg$open,1,2,1)}. 
This result triggers postdiction rule (\ref{lst:lp-effCondTrans3}) and knowledge about an abnormality is produced: \co{knows(ab\_open,0,2,1)}. 
Consequently, the wheelchair has to follow another route to achieve the goal. For branch 0, we have \co{knows(open,1,2,0)} after the sensing.
This result triggers the postdiction rule (\ref{lst:lp-effCondTrans2}):
Because \co{knows($\neg$open,0,2,0)} and \co{knows(open,1,2,0)} hold, one can postdict that there was no abnormality when \co{open} occurred: \co{knows($\neg$ab\_open,0,2,0)}.
Finally, the robot can drive through the door: \co{occ(drive,2,0)} and the causation rule (\ref{lst:lp-effCondTrans1}) triggers knowledge that the robot is in the living room at step 3:  \co{knows(in\_liv,3,3,0)}.
\end{sloppypar}

\vspace{-10pt}
\section{Conclusion}\label{sec:conclusion}
\vspace{-10pt}
We developed an approximation of the possible worlds semantics with elaboration tolerant support for postdiction, and implemented a planning system by a translation of the approximation to ASP. 
We show that the plan existence problem in our framework can be solved in NP.
We relate our approach to the \pws\ semantics of \Ak\ by extending \Ak\ semantics to allow for temporal queries. 
We show that \hpx\ is sound \wrt this semantics. 
Finally, we provide a proof of concept for our approach with the case study in Section \ref{sec:evalAndCasStudy}. 
An extended version of the Case Study will appear in \citep{Eppe2013a}.
Further testing revealed the inferiority of the \hpx\ implementation to dedicated PDDL planners like CFF \citep{Hoffmann2005}. 
This result demands future research concerning the transfer of heuristics used in PDDL-based planners to ASP.


{\small
\bibliographystyle{abbrvnat}
\bibliography{bib}
}

\end{document}

%% file: macros.tex
\usepackage{listings}
\usepackage{array}
\usepackage[ruled]{algorithm}
\usepackage[noend]{algpseudocode} 
\usepackage{todonotes}
\usepackage{rotating}
\usepackage{color}
\usepackage{amsmath}
\usepackage{amssymb}

\usepackage{paralist}
\usepackage{enumitem}
\usepackage{nccmath}

\newcommand{\code}[1]{{\texttt{\footnotesize #1}}}
\newcommand{\co}[1]{\code{#1}}
\renewcommand{\cos}[1]{\codes{#1}}
\newcommand{\codes}[1]{{\texttt{\scriptsize #1}}}

\renewcommand{\i}[1]{\textit{#1}}
\newcommand{\f}[1]{\textbf{#1}}

\newcommand{\itz}[1]{\begin{itemize} 
#1 \end{itemize}}

\newcommand{\ienum}[1]{\begin{inparaenum}[\itshape (a\upshape)]
#1 \end{inparaenum}}

\newcommand{\enum}[1]{\begin{enumerate}
#1 \end{enumerate}}
\newcommand {\hpx}[0]{\mc{HPX}}

\newcommand{\opn}[1]{\operatorname{#1}}
\newcommand{\symbn}[1]{\opn{#1}}
\newcommand{\sn}[1]{\symbn{#1}}
\newcommand{\sns}[1]{\text{{\rmfamily{\scriptsize{#1}}}}}

\newcommand{\rab}[1]{\left\langle #1 \right\rangle}

\newcommand{\ol}[1]{\overline{#1}}

\newcommand{\eg}[0]{e.g.~}
\newcommand{\ie}[0]{i.e.~}
\newcommand{\wrt}[0]{wrt.~}

\lstdefinestyle{pddl}{
		basicstyle=\ttfamily\small,
		stepnumber=1, 
		numberstyle=\small, 
		numbersep=10pt,
}
\lstdefinestyle{lp}{
		basicstyle=\ttfamily\small\color{black},
		numbers=left, 
		numberstyle=\small, 
		numbersep=10pt,
		xleftmargin=2em,
		morecomment=[l][\color{black}]{\%},
}

\newcommand{\mc}[1]{\ensuremath{\mathcal{#1}}}

\newcommand{\ms}[0]{\hspace{3pt}}
\newcommand{\lam}[0]{\leftarrow}{}
\definecolor{gray}{gray}{.5}  

\algrenewcommand\algorithmicthen{}
\algrenewcommand\algorithmicdo{}

\newcommand{\pws}[0]{\mc{PWS}}
\newcommand{\Ak}[0]{$\mc{A}_k$}
\newcommand{\Ack}[0]{$\mc{A}_k^c$}

\newcommand{\A}[0]{$\mc{A}$}

\lstnewenvironment{pddl}[1]{}{}
\newcommand{\li}[1]{\i{l.}~\ref{#1}}
\newcommand{\lir}[2]{\i{ll.}~\ref{#1}-\ref{#2}}
\newcommand{\Li}[1]{Line~\ref{#1}}
\newcommand{\Lir}[2]{Lines~\ref{#1}-\ref{#2}}

\newcommand{\hide}[1]{}